# Mask-PINNs: Mitigating Internal Covariate Shift in Physics-Informed Neural Networks

Feilong Jiang[a] Xiaonan Hou[a] Jianqiao Ye[a] Min Xia[b]*
[a] Department of Engineering, Lancaster University, LA1 4YW Lancaster, U.K.
[b] Department of Mechanical and Materials Engineering, University of Western Ontario, London, Ontario, Canada
min.xia@uwo.ca

## Abstract

Physics-Informed Neural Networks (PINNs) have emerged as a powerful framework for solving partial differential equations (PDEs) by embedding physical laws directly into the loss function. However, as a fundamental optimization issue, internal covariate shift (ICS) hinders the stable and effective training of PINNs by disrupting feature distributions and limiting model expressiveness. Unlike standard deep learning tasks, conventional remedies for ICS—such as Batch Normalization and Layer Normalization—are not directly applicable to PINNs, as they distort the physical consistency required for reliable PDE solutions. To address this issue, we propose Mask-PINNs, a novel architecture that introduces a learnable mask function to regulate feature distributions while preserving the underlying physical constraints of PINNs. We provide a theoretical analysis showing that the mask suppresses the expansion of feature representations through a carefully designed modulation mechanism. Empirically, we validate the method on multiple PDE benchmarks—including convection, wave propagation, and Helmholtz equations—across diverse activation functions. Our results show consistent improvements in prediction accuracy, convergence stability, and robustness. Furthermore, we demonstrate that Mask-PINNs enable the effective use of wider networks, overcoming a key limitation in existing PINN frameworks.

Key words: Deep learning, Physics-informed neural networks, Partial differential equations, Scientific machine learning

## 1 Introduction

As a bridging approach between classical scientific computing and machine learning, Physics-Informed Neural Networks (PINNs) have emerged as a powerful framework for solving partial differential equations (PDEs) [1]. In PINNs, neural networks approximate the solution of a PDE while the governing equations are enforced through the loss function. This approach enables data-efficient learning and provides a flexible framework for incorporating prior physical knowledge [2-5]. Due to their strong interpretability and ease of implementation, PINNs have found successful applications in diverse fields such as fluid mechanics [6], heat transfer problems [7], solid mechanics [8], geoengineering [9], chemical engineering [10], finance [11], and climate modelling [12].

Despite their growing popularity, training PINNs effectively remains challenging. Numerous studies have been conducted to improve their performance. Most of these methods focus on training pathologies of PINNs. To mitigate activation sensitivity, several methods have been proposed to learn or adapt activation functions during training thereby improve efficiency and stability of training [13], [14]. To address the performance degradation in deeper PINN architectures, methods like PirateNets, HyResPINNs, and Deeper-PINNs are proposed to improve the scalability of PINNs [15-17]. To overcome vanishing or exploding gradients caused by poor initial choices, theoretically informed initialization approaches are proposed to ensure stable training and robust convergence in PINNs [18], [19]. Another well-known challenge comes from the composite loss function. Different loss components (e.g., PDE residuals, boundary conditions, data terms) often compete, leading to gradient conflicts, which hinder convergence. Numerous methods have been proposed to alleviate such imbalanced optimization [20-23]. Moreover, as demonstrated in [24-26], adopting more effective network architectures can strengthen the representational capacity of PINNs, offering a promising route to improve the performance of PINNs.

One critical yet underexplored challenge in PINNs is internal covariate shift—the instability of feature distributions during training—which undermines both the learning dynamics and the expressiveness of PINNs [27]. While normalization techniques like Batch Normalization (BN) and Layer Normalization (LN) are common in deep learning to mitigate this, their reliance on batch or feature statistics can disrupt the input-output mapping and the relative relationships between different inputs. This poses a fundamental question: How can we alleviate the internal covariate shift (ICS) of PINNs while preserving their physical consistency? In this paper, we introduce Mask-PINNs, a novel PINN architecture designed to address this issue. Our contributions are summarized as follows:

- We propose Mask-PINNs, a novel PINN architecture that introduces a learnable, nonlinear mask function to regulate the internal feature distributions. Unlike conventional normalization techniques, the mask function operates in a pointwise and input-conditioned manner, ensuring that deterministic input–output relationships are preserved — a critical requirement for PINNs.
- We demonstrate that the proposed mask mechanism effectively suppresses feature drift, leading to more centered and stable feature distribution throughout training. This allows the model to maintain expressiveness without sacrificing physical fidelity.
- Extensive experiments across diverse PDE benchmarks and activation functions show that Mask-PINNs achieve consistent and significant improvements in prediction accuracy and training stability.
- We show that Mask-PINNs enable effective training of wider neural networks, overcoming a critical limitation in conventional PINNs where increasing network width typically leads to performance degradation. This makes wider networks viable in practice, expanding the design space for PINNs.

The remainder of the paper is organized as follows: Section 2 reviews relevant prior work and discusses key limitations; Section 3 presents Mask-PINNs and shows theoretically how the mask improves stability and signal propagation; Section 4 presents comprehensive experimental validation; Section 5 analyzes the Neural Tangent Kernel of Mask-PINNs to provide further theoretical insight into their training dynamics; and finally, Section 6 summarizes the findings and discusses potential directions for future research.

# 2 Problem Statement

This section introduces the working principle of PINNs. We also discuss traditional normalization techniques' limitations in PINNs.

## 2.1 Physics-Informed Neural Networks

Given the following partial differential equation (PDE):

$$\mathcal{N}\left[\mathbf{u}(\mathbf{x})\right]=f(\mathbf{x}),\ \mathbf{x}\in\Omega, \tag{1}$$

$$\mathcal{B}\left[\mathbf{u}(\mathbf{x})\right]=g(\mathbf{x}),\ \mathbf{x}\in\partial\Omega, \tag{2}$$

where $\mathbf{u}(\mathbf{x})$ indicates the solution, $\mathcal{N}[\cdot]$ represents a differential operator, $\mathcal{B}[\cdot]$ indicates boundary operator.

The task of PINNs is approximating the solution of the given PDE by a deep neural network $\mathbf{u}_\theta(\mathbf{x})$, where $\theta$ denotes all trainable parameters. For PINNs, $\mathbf{u}_\theta(\mathbf{x})$ is constrained by the following loss function:

$$\mathcal{L}(\theta) = \lambda_b \mathcal{L}_b(\theta) + \lambda_r \mathcal{L}_r(\theta), \tag{3}$$

where

$$\mathcal{L}_b(\theta) = \frac{1}{N_b} \sum_{i=1}^{N_c} \left| \mathcal{B}\left[\mathbf{u}_\theta\left(\mathbf{x}_b^i\right)\right] - g\left(\mathbf{x}_b^i\right) \right|^2, \tag{4}$$

$$\mathcal{L}_r(\theta) = \frac{1}{N_r} \sum_{i=1}^{N_r} \left| \mathcal{N}\left[\mathbf{u}_\theta\left(\mathbf{x}_r^i\right)\right] - f\left(\mathbf{x}_r^i\right) \right|^2, \tag{5}$$

$\lambda$ is responsible for balancing the weight of different loss terms. $\mathbf{x}_r$ are collocation points in the domain. $\mathbf{x}_b$ are boundary points.

## 2.2 Why Standard Normalization Fails in PINNs

During training, the feature distributions at each layer can drift in mean, variance and overall shape, a phenomenon referred to as internal covariate shift. For bounded activation functions such as Tanh, such drift can drive outputs toward their saturation limits, leading to vanishing gradients and reduced expressiveness [28]. Although unbounded activations such as GELU, SiLU, or SoftPlus can alleviate this effect, they remain susceptible to ICS and often behave quasi-linearly when inputs are far from zero, limiting the network's ability to capture complex nonlinear patterns [29], [30].

In conventional deep learning, ICS is typically addressed by normalization techniques such as batch normalization BN or LN [27], [31], [32]. However, in the context of PINNs, traditional these methods can distort the underlying physical relationships between inputs and outputs, due to their reliance on batch-level or feature-level statistics.

To clarify this issue, we briefly revisit the formulations of BN and LN:

The output of a BN layer can be written as:

$$\hat{h}(\mathbf{x}_i;\mathcal{S}) = \gamma\left(\frac{h(\mathbf{x}_i) - \mu_B}{\sigma_B + \varepsilon}\right) + \beta, \tag{6}$$

where $h(\mathbf{x}_i)$ is the pre-activation feature map for input $\mathbf{x}_i$, $\mathcal{S}$ is the current batch contains $\mathbf{x}_i$, $\varepsilon$ is for numerical stability in case the denominator becomes zero by chance. $\gamma$ and $\beta$ are learnable scale and shift.

$$\mu_B = \frac{1}{N} \sum\nolimits_{x_j \in \mathcal{S}} h(\mathbf{x}_j), \tag{7}$$

$$\sigma_B^2 = \frac{1}{N} \sum\nolimits_{\mathbf{x}_j \in \mathcal{S}} \left(h(\mathbf{x}_j) - \mu_B\right)^2, \tag{8}$$

where $N$ denotes the batch size.

As $\mu_B$ and $\sigma_B$ are computed using all samples in the batch, the normalized output $\hat{h}(\mathbf{x}_i;\mathcal{S})$ is no longer solely a function of $\mathbf{x}_i$, but also implicitly depends on every $\mathbf{x}_j \in \mathcal{S}$. Consequently, the final output can be expressed as:

$$u(\mathbf{x}_i) = f\left(\mathbf{x}_i; \{\mathbf{x}_j\}_{j \in \mathcal{S}}\right). \tag{9}$$

The output $u(\mathbf{x}_i)$ now depends on the entire batch. Since PINNs enforce PDE constraints via input derivatives, the batch dependence of $u(\mathbf{x}_i)$ implies that its derivatives are influenced by unrelated samples, thereby violating the requirement of pointwise computation.
Given an input $\mathbf{x}_i$, and $h(\mathbf{x}_i) \in \mathbb{R}^d$ be the pre-activation output of an FC layer. LN produces:

$$\hat{h}(\mathbf{x}_i) = \gamma \frac{h(\mathbf{x}_i) - \mu_i}{\sigma_i + \varepsilon} + \beta, \tag{10}$$

where $\mu_i = \frac{1}{d}\sum_{j=1}^{d} h_j(\mathbf{x}_j)$, $\sigma_i^2 = \frac{1}{d}\sum_{j=1}^{d}\left(h_j(\mathbf{x}_i) - \mu_i\right)^2$. For any input $\mathbf{x}_i$, the normalized vector $\frac{h(\mathbf{x}_i) - \mu_i}{\sigma_i}$ is zero-mean and unit-variance.

LN thus enforces identical statistical properties (mean and variance) across features for every input, regardless of their physical differences. In PINNs, where inputs carry specific physical meaning, uniform normalization can suppress or distort meaningful physical variations.
In the following, we empirically confirm these issues with an experiment on the heat conduction problem:

$$\frac{\partial u}{\partial t} - \frac{\partial^2 u}{\partial x^2} = 0, \ (x,t) \in [0,1] \times [0,1], \tag{11}$$

$$u(x,0) = \sin(\pi x), \ x \in [0,1], \tag{12}$$

$$u(0,t) = u(1,t) = 0, \ t \in [0,1], \tag{13}$$

the analytical solution for this problem is:

$$u(x,t) = \sin(\pi x) \cdot \exp(-t\pi^2). \tag{14}$$

All the outcomes are averaged from 5 independent trials. As shown in Fig. 1, BN-based PINN (BN-PINN) and LN-based PINN (LN-PINN) show worse results than Vanilla PINN. For BN-PINN, even though the loss curve decreases during the training, the relative $L^2$ error curve almost stays flat. This indicates that for PINN with BN, the model is not trained with the given physical constraint. Although LN-PINN shows a consistent decrease in both loss and relative $L^2$ error, the accuracy is lower than that of vanilla method. This indicates that LN impairs the model's ability to capture physical details.

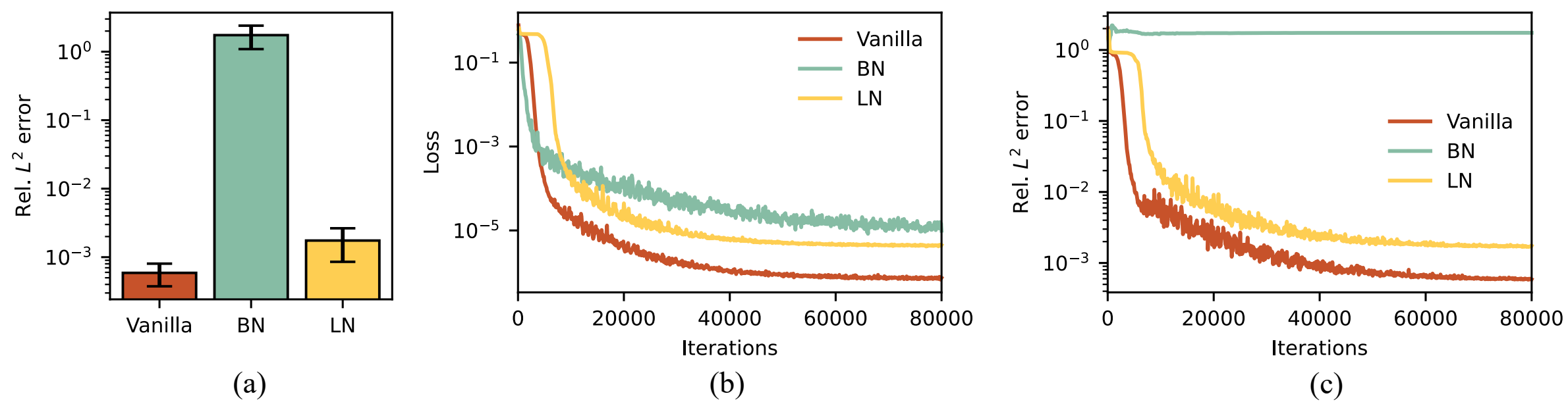

Figure 1. Performance comparison of Vanilla PINN, BN-PINN, and LN-PINN. (a) Mean relative $L^2$ error. (b) Averaged loss curve. (c) Averaged relative $L^2$ error curve.

# 3 Methodology

## 3.1 Motivation and Design Principles

The motivation behind the proposed method is to reduce ICS in PINNs while preserving the physical consistency inherent in the input-output relationships. Instead of forcing the feature representations to conform to fixed statistical distributions, as BN and LN do, we propose an input-conditioned modulation mechanism. Fig. 2 shows the schematic illustration of the proposed method. The proposed approach leverages a tailored mask function to control and stabilize the output distributions across hidden layers.

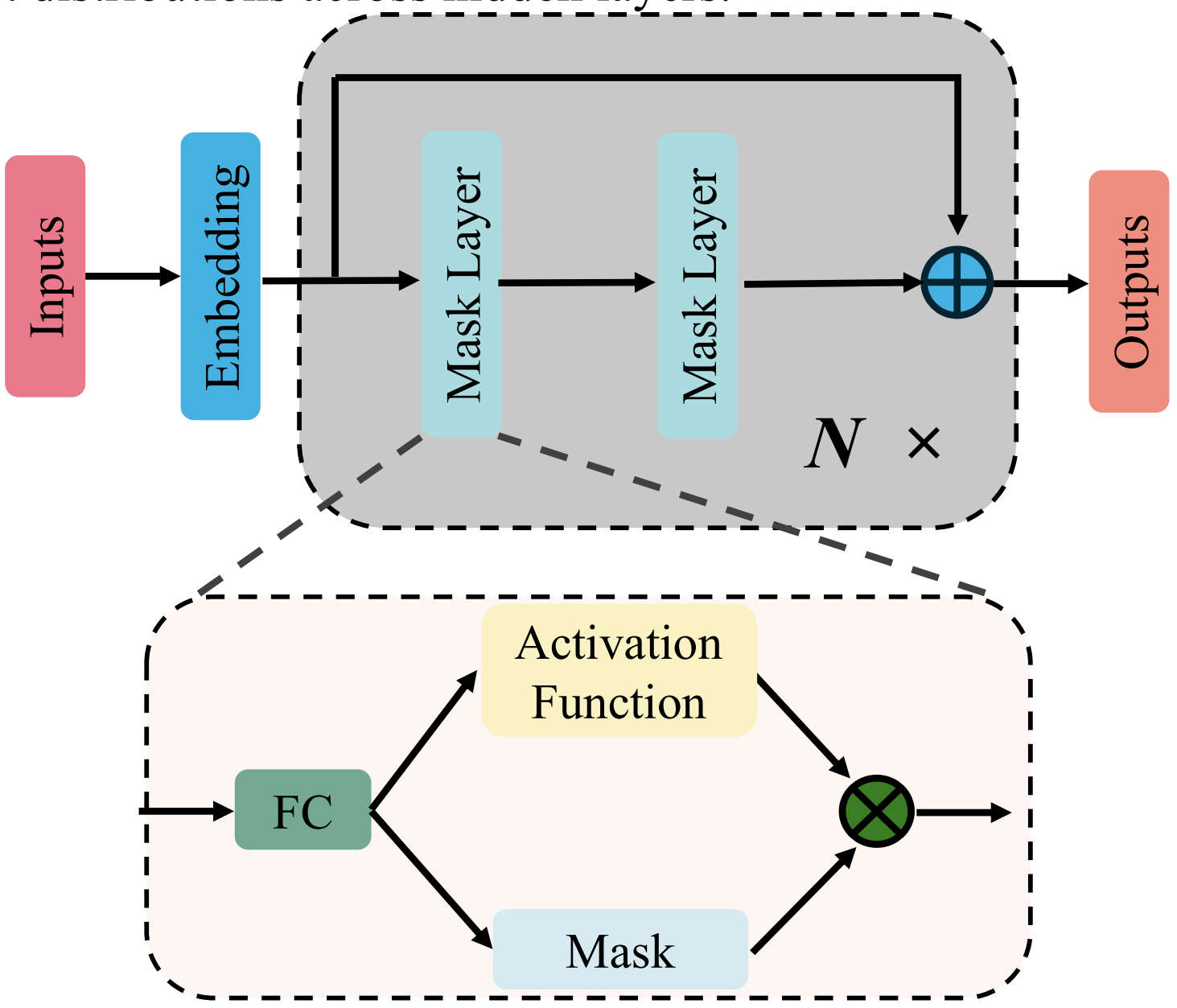

Figure 2. Illustration of the architecture of the Mask-PINNs.

Given an input vector $\mathbf{x}$, a mask function can be expressed as:

$$F(\mathbf{x}) = 1 - \exp\left[-(\boldsymbol{\alpha} \odot \mathbf{x})^2\right] \tag{15}$$

where $\boldsymbol{\alpha}$ is a learnable vector that controls the sharpness of the mask function. $\boldsymbol{\alpha}$ is initialized as a positive vector. Fig. 3 shows the visualization of the proposed mask function. The proposed mask function, resembling an inverted Gaussian centered at zero, is smooth and differentiable, ensuring the network output remains differentiable as required by PINNs. Moreover, it modulates each sample independently, thereby maintaining the distinctiveness of individual inputs.

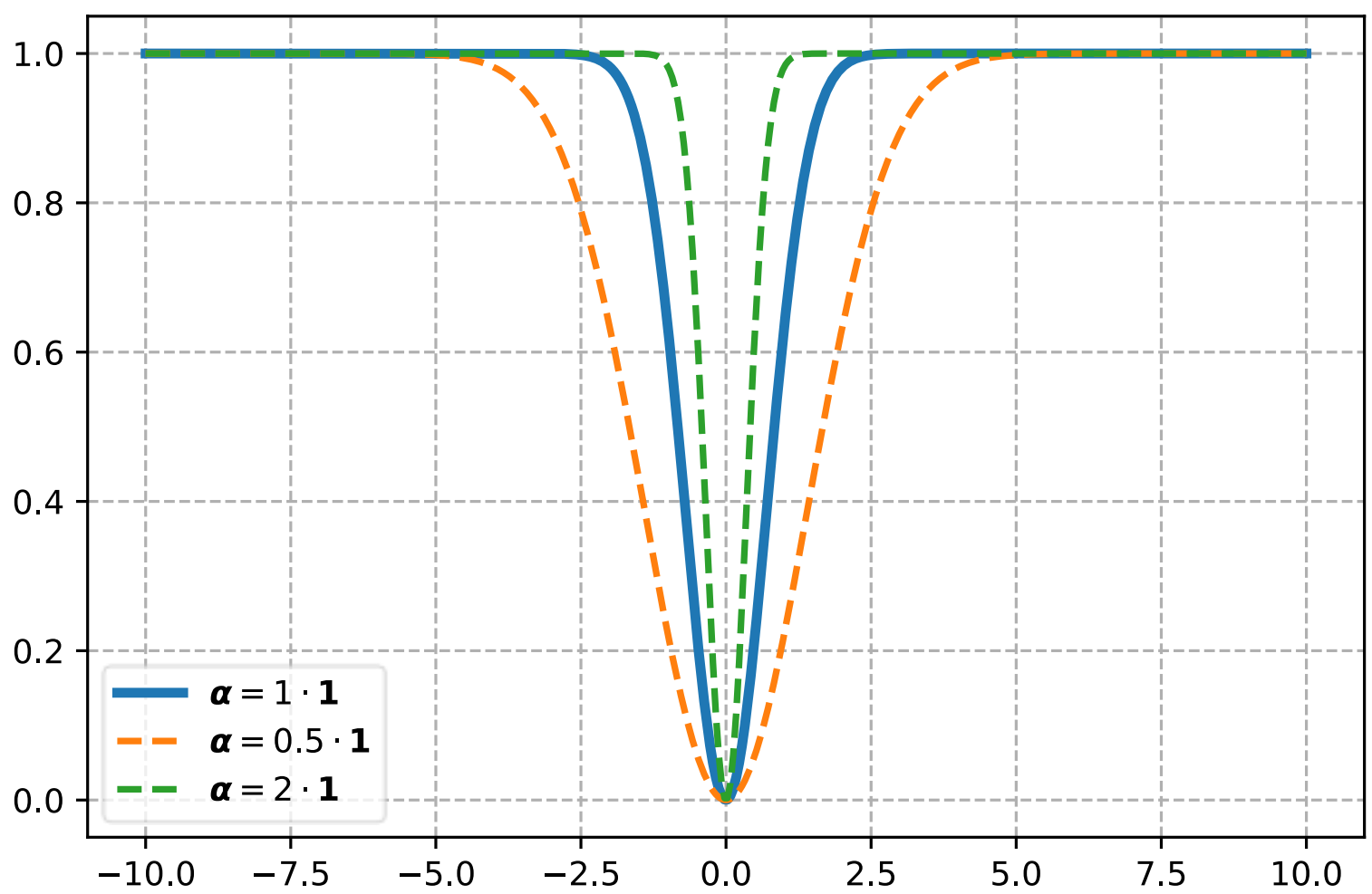

Figure 3. Mask function with different $\boldsymbol{\alpha}$.

With the mask inserted, the feed-forward process of layer $l$ can be expressed as:

$$z^{(l)} = w^{(l)} H + b^{(l)}, \tag{16}$$

$$H^{(l)} = F^{(l)}(z^{(l)}) \odot \sigma(z^{(l)}), \tag{17}$$

where $z$ represents the pre-activation, $l$ is the number of the layer, $H$ denotes the output of a layer, $w$ represents weights and $b$ represents biases.

### 3.2 Theoretical Analysis

**Remark 1.** For any scalar input $u \in \mathbb{R}$, since $F(u) \le 1$, we have:

$$|H(u)| \le |\sigma(u)|. \tag{18}$$

Thus, the mask contracts activation magnitudes globally, suppressing extreme values and yielding a more compact feature distribution.

**Proposition 1.** For any $r > 0$, quantify local sensitivity by the minimal Lipschitz constant on $[-r, r]$:

$$L_H(r) := \sup_{x \ne y,\ |x|,|y| \le r} \frac{|H(x) - H(y)|}{|x - y|},\ L_\sigma(r) := \sup_{x \ne y,\ |x|,|y| \le r} \frac{|\sigma(x) - \sigma(y)|}{|x - y|}. \tag{19}$$

For differentiable scalar functions on an interval, the mean value theorem implies

$$L_H(r) := \sup_{|u| \le r} |H'(u)|,\ L_\sigma(r) := \sup_{|u| \le r} |\sigma'(u)|. \tag{20}$$

As $r \to 0$

$$\frac{L_H(r)}{L_\sigma(r)} = \begin{cases} O(r) \text{ if } \sigma(0) \ne 0, \\ O(r^2) \text{ if } \sigma(0) = 0. \end{cases} \tag{21}$$

*Proof.* Using the local expansions near 0:

$$F(u) = (\alpha u)^2 + O(u^4), \tag{22}$$

$$F'(u) = 2\alpha^2 u + O(u^3). \tag{23}$$

$$\sigma(u) = \sigma(0) + \sigma'(0)u + O(u^2), \tag{24}$$

$$\sigma'(u) = \sigma'(0) + O(u). \tag{25}$$

We have:

$$H'(u) = F'(u)\sigma(u) + F(u)\sigma'(u). \tag{26}$$

If $\sigma(0) = 0$:

$$H'(u) = 3\alpha^2 \sigma'(0) u^2 + O(u^3). \tag{27}$$

Hence:

$$L_H(r) = 3\alpha^2 |\sigma'(0)| r^2 + O(r^3). \tag{28}$$

If $\sigma(0) \ne 0$:

$$H'(u) = 2\alpha^2 \sigma(0) u + O(u^2). \tag{29}$$

Hence:

$$L_H(r) = 2\alpha^2 |\sigma(0)| r + O(r^2). \tag{30}$$

For vanilla method:

$$L_\sigma(r) = |\sigma'(0)| + O(r). \tag{31}$$

Dividing the leading asymptotics gives the stated result.

**Remark 2.** Under Xavier initialization, pre-activations typically concentrate near zero. In this regime, Proposition 2 shows that the proposed mask contracts the local Lipschitz constant by at least one order compared to the vanilla method. This contraction dampens the effect of perturbations, thereby keeping the output zero cantered and stable around initialization.

In summary, the proposed mask mechanism globally contracts activation magnitudes, suppressing extreme values and keeping feature distributions compact, while it locally stabilizes behavior near initialization by reducing sensitivity to perturbations. These properties act jointly to mitigate ICS and promote reliable signal propagation in PINNs.

### 3.3 Mask Placement

Notice that unlike normalization methods, which are typically applied before the activation function, our mask function is applied alongside it. Such design helps the mask function robustly scale down the feature values across different activation functions.

Suppose placing the mask function before the activation function, we have the output as:

$$H = \sigma(z \odot F(z)). \tag{32}$$

Although this may seem to be a more straightforward approach, the distribution of $H$ depends heavily on the properties of the activation function, which can therefore reintroduce the ICS. For activations with $\sigma(0) \neq 0$ (e.g., SoftPlus), such design leads to biased outputs even for zero-centered inputs, causing a systematic shift in feature distributions across layers. In contrast, our theoretical analysis shows that applying the mask alongside the activation function can effectively suppresses the activation magnitude while alleviating the mean shift caused by activation functions.

## 4 Experiments

To demonstrate the effectiveness of the proposed method, we conducted experiments across several benchmarks. The methods compared against the proposed methods include Vanilla PINNs [1], ResNet-based PINNs (ResNet-PINN), Weight Normalization-based PINNs (WN-PINNs), locally adaptive activation functions (LAAF) based PINNs [33], PirateNets [15], and [13]. All the methods are trained with Tanh, GELU, SiLU, and SoftPlus as activation functions. Note that ABU-PINN is specifically designed to automatically select activation functions during training, and therefore produces only one result, while the other methods are evaluated separately with each of the four activation functions. Due to the difference in the structure, the depth of each model may vary, but the total trainable parameters are kept about the same. Adam is utilized as the optimizer. All experiments are conducted on a single NVIDIA GeForce RTX 4090 GPU. All results are averaged over 5 independent trials.

### 4.1 Convection Equation

We consider convection equation expressed as below:

$$\frac{\partial u}{\partial t} + \beta \frac{\partial u}{\partial x} = 0,\ x \in [0,\ 2\pi],\ t \in [0,\ 1], \tag{33}$$

$$u(0,\ x) = \sin(x), \tag{34}$$

$$u(t,\ 0) = u(t,\ 2\pi), \tag{35}$$

where we set $\beta$ as 30.

Each model is composed of 13 hidden layers, each containing 256 neurons, and is trained for 50,000 iterations. For the proposed method, $\boldsymbol{\alpha}$ is initialized with all entries set to 1.0.

Table 1 shows the outcomes of different models. The best-performing method under each activation function is underlined, while the overall best result across all methods is highlighted in bold. As can be seen, the proposed method achieves consistent best result across different activation functions. Specifically, for Tanh, the proposed method achieves performance a relative $L^2$ error 1 order of magnitude lower than that of vanilla method. For SoftPlus, the result of the proposed method is 2 orders of magnitude lower than vanilla method.

Table 1: Convection equation: relative $L^2$ error of different models.

| Method | Tanh | GELU | SiLU | SoftPlus |
|---|---|---|---|---|
| Vanilla | 7.29e-3 | 9.91e-4 | 1.37e-3 | 3.55e-1 |
| ResNet-PINN | 1.75e-2 | 1.74e-3 | 1.50e-3 | 2.41e-2 |
| LAAF | 7.30e-3 | 9.44e-4 | 7.23e-4 | 7.35e-2 |
| PirateNet | 1.98e-3 | 9.73e-3 | 4.54e-3 | 1.39e-2 |
| WN-PINNs | 3.17e-3 | 1.02e-3 | 4.89e-3 | 5.12e-3 |
| ABU-PINN | 2.98e-3 | | | |
| Mask-PINN | <u>6.93e-4</u> | <u>5.26e-4</u> | **<u>4.79e-4</u>** | <u>1.66e-3</u> |

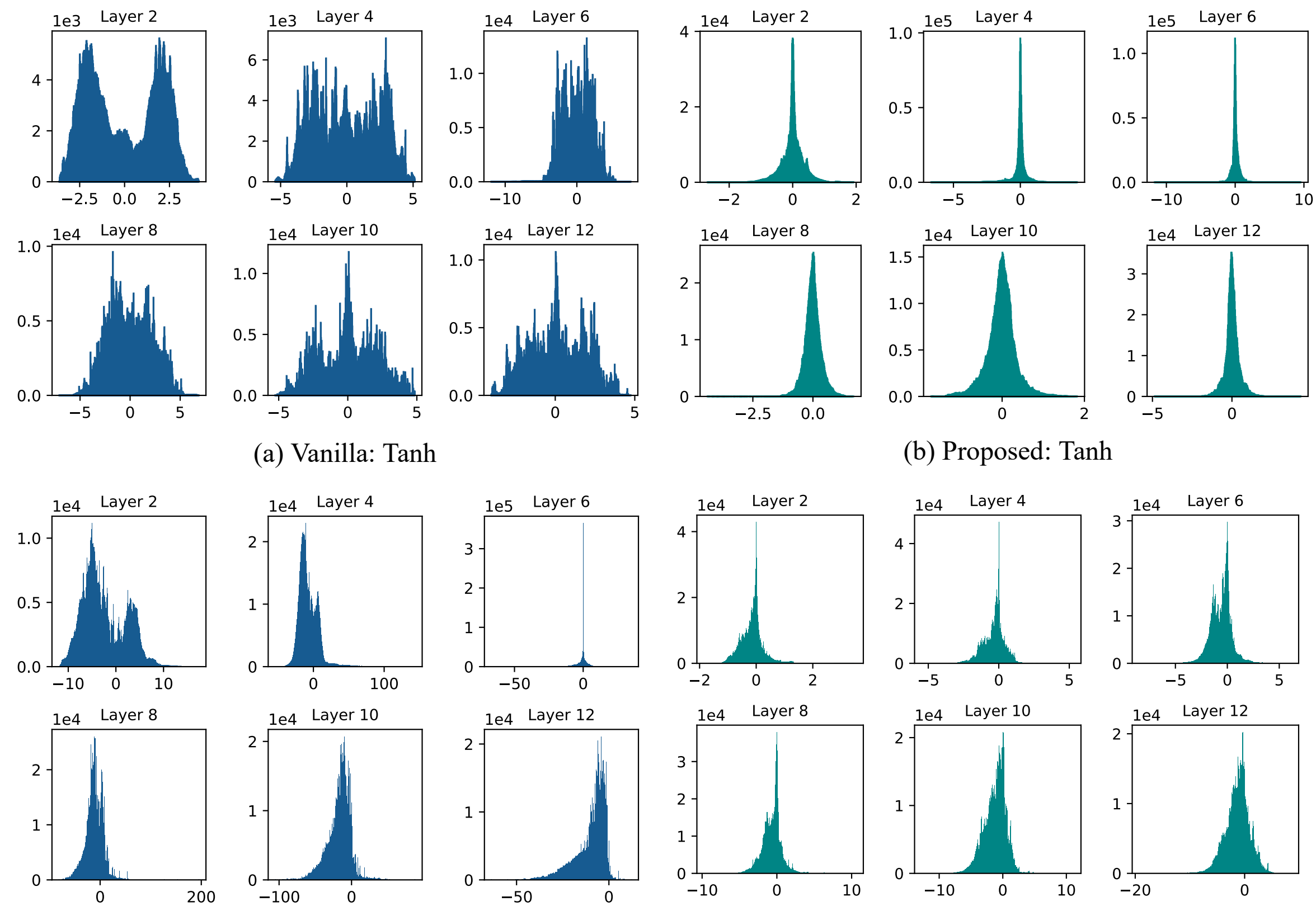

(a) Vanilla: Tanh (b) Proposed: Tanh

(c) Vanilla: SoftPlus (d) Proposed : SoftPlus

Figure 4. Convection Equation: pre-activation distributions of the vanilla and proposed methods.

Fig. 4 gives the pre-activation distributions of the vanilla and proposed methods at the end of training, using SoftPlus and Tanh as activation function. For Tanh, most of the vanilla pre-activations drift away from the origin, pushing many neurons into the saturated regime where gradients vanish. However, the proposed method maintains a more uniform, Gaussian-like distribution across hidden layers, which prevents saturation and ensures stable gradient flow throughout the network. This distributional stability accounts for the superior performance of the proposed method over the vanilla baseline. For SoftPlus, as the layer goes deeper, the spread of the feature values changes dramatically, and more values shift to the negative side. This arises from the fact that SoftPlus is a unsymmetric activation function with $\mathrm{SoftPlus}(0) \neq 0$, which introduces a systematic bias in the feature values, as discussed in Section 3.3. Besides, such phenomenon implies that during training, the vanilla method's pre-activations tend to drift toward the linear regime of the SoftPlus activation function, which can negatively impact both the training dynamics and the model's expressiveness. In contrast, the proposed method shows a much slower change in the scale change and keeps the feature distributions in a Gaussian-like form. This distinct difference between the feature distributions explains why our method yields significantly better performance when using SoftPlus.

### 4.2 1D Wave Propagation

Here we consider the 1D wave propagation equation as:

$$\frac{\partial^2 u}{\partial t^2} = c^2 \frac{\partial^2 u}{\partial x^2},\ (x,t) \in [0,1] \times [0,5], \tag{36}$$

$$u(0,t) = u(1,t) = 0,\ t \in [0,5], \tag{37}$$

$$u(x,0) = \sin(\pi x),\ x \in [0,1], \tag{38}$$

$$\frac{\partial u}{\partial t}(x,0) = 0, \tag{39}$$

the analytical solution for this problem is:

$$u(x,t) = \sin(\pi x)\cos(c\pi t). \tag{40}$$

Here we set $c = 1$. All the models have 7 hidden layers, each containing 256 neurons, and is trained for 20,000 iterations. For the proposed method, $\boldsymbol{\alpha}$ is initialized with all entries set to 1.0.

Table 2: 1D Wave propagation: relative $L^2$ errors of different models.

| Method | Tanh | GELU | SiLU | SoftPlus |
|---|---|---|---|---|
| Vanilla | 7.33e-2 | 5.09e-3 | 8.52e-3 | 3.61e-2 |
| ResNet-PINN | 4.85e-2 | 7.52e-3 | 5.22e-3 | 7.67e-1 |
| LAAF | 1.04e-2 | 4.37e-3 | 5.73e-3 | 1.31e-2 |
| PirateNet | 1.59e-2 | 8.83e-3 | 9.95e-3 | 5.04e-2 |
| WN-PINNs | 6.14e-2 | 5.60e-3 | 7.30e-3 | 1.46e-1 |
| ABU-PINN | | 4.82e-2 | | |
| Mask-PINN | **2.72e-3** | 2.90e-3 | 4.82e-3 | 6.38e-3 |

Table 2 shows the outcomes of different models. The proposed method consistently achieves best result across different activation functions. Interestingly, while the vanilla method yielded the poorest performance with Tanh, the proposed method achieved its best result using the same activation function. In Fig. 5, we show the variance evolution of the proposed method and vanilla method with Tanh as activation function. The variance for vanilla method increases notably with training, suggesting severe distribution drift. The variance curves of the proposed method show relatively smaller values and stable trends after initial iterations, indicating controlled variance growth. This reflects effective regulation of pre-activations as depth and training progress.

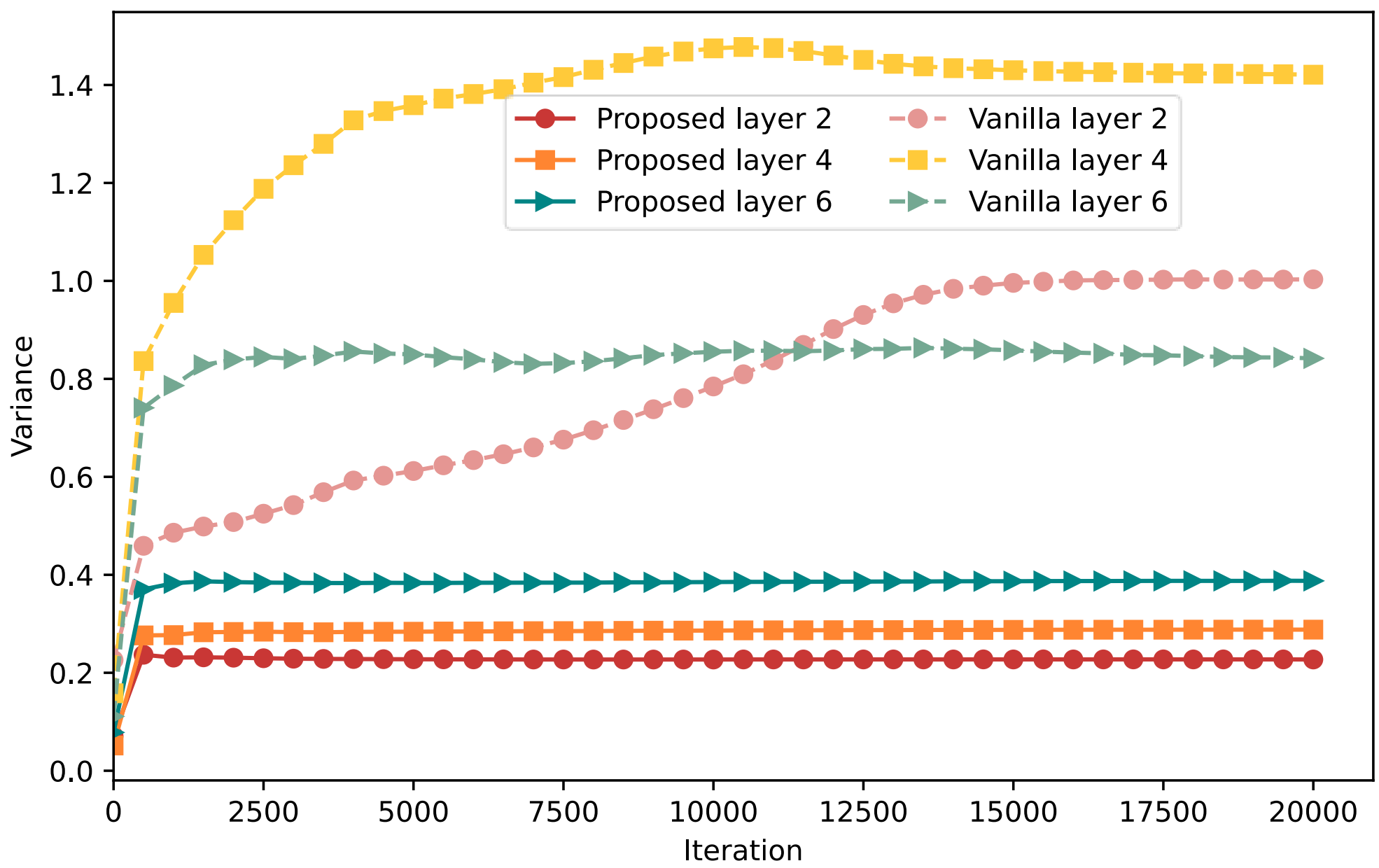

Figure 5. 1D wave propagation equation: variances of pre-activations of vanilla and proposed methods with Tanh activation.

### 4.3 Helmholtz Equation

We consider the Helmholtz equation expressed as below:

$$\frac{\partial^2 u}{\partial x^2}+\frac{\partial^2 u}{\partial y^2}+k^2 u-q(x,y)=0,\ x\in[-1,\ 1],\ y\in[-1,\ 1], \tag{41}$$

$$u(-1,y)=u(1,y)=u(x,-1)=u(x,1)=0, \tag{42}$$

where the source term $q(x,y)$ is in the form of:

$$\begin{aligned} q(x,y)=&-(a_1\pi)\sin(a_1\pi x)\sin(a_2\pi y)\\ &-(a_2\pi)\sin(a_1\pi x)\sin(a_2\pi y)\\ &+k^2\sin(a_1\pi x)\sin(a_2\pi y), \end{aligned} \tag{43}$$

here we set $a_1=6$, $a_2=6$ and $k=1$.

All models are composed of 11 hidden layers with 128 neurons each, except PirateNet, which uses 10 hidden layers to keep about the same number of parameters with other methods. Each model is trained for 50,000 iterations. For the proposed method, $\boldsymbol{\alpha}$ is initialized with all entries set to 1.0 for GELU and SiLU, 5.0 for Tanh, and 2.0 for SoftPlus.

Table 3: Helmholtz equation: relative $L^2$ errors of different models.

| Method | Tanh | GELU | SiLU | SoftPlus |
|---|---|---|---|---|
| Vanilla | 9.37e-1 | 7.20e-1 | 1.00 | 1.00 |
| ResNet-PINN | 3.48e-1 | 8.02e-1 | 7.10e-1 | 1.00 |
| LAAF | 3.26e-2 | 2.26e-2 | 1.93e-1 | 5.89e-1 |
| PirateNet | 1.17 | 3.56e-2 | 2.68e-1 | 7.15e-1 |
| WN-PINNs | 7.70e-1 | 7.12e-1 | 7.12e-1 | 1.00 |
| ABU-PINN | 2.83e-2 | | | |
| Mask-PINN | **9.94e-3** | 1.52e-2 | 1.99e-2 | 1.75e-2 |

Table 3 shows the outcomes of different models. As can be seen, the proposed method achieves the best performance across different activation functions. Notably, for activation functions SiLU and SoftPlus, other methods fail to get reliable results. In contrast, our model maintains comparable accuracy across all activation functions. This highlights its robustness across different activation functions.
Since the vanilla method performs best with GELU, we report the averaged loss curves of all models under this activation in Fig. 6 (ABU-PINN is excluded as it dynamically selects activation functions during training.) As shown, our method converges faster and reaches a lower final loss, indicating improved training dynamics.

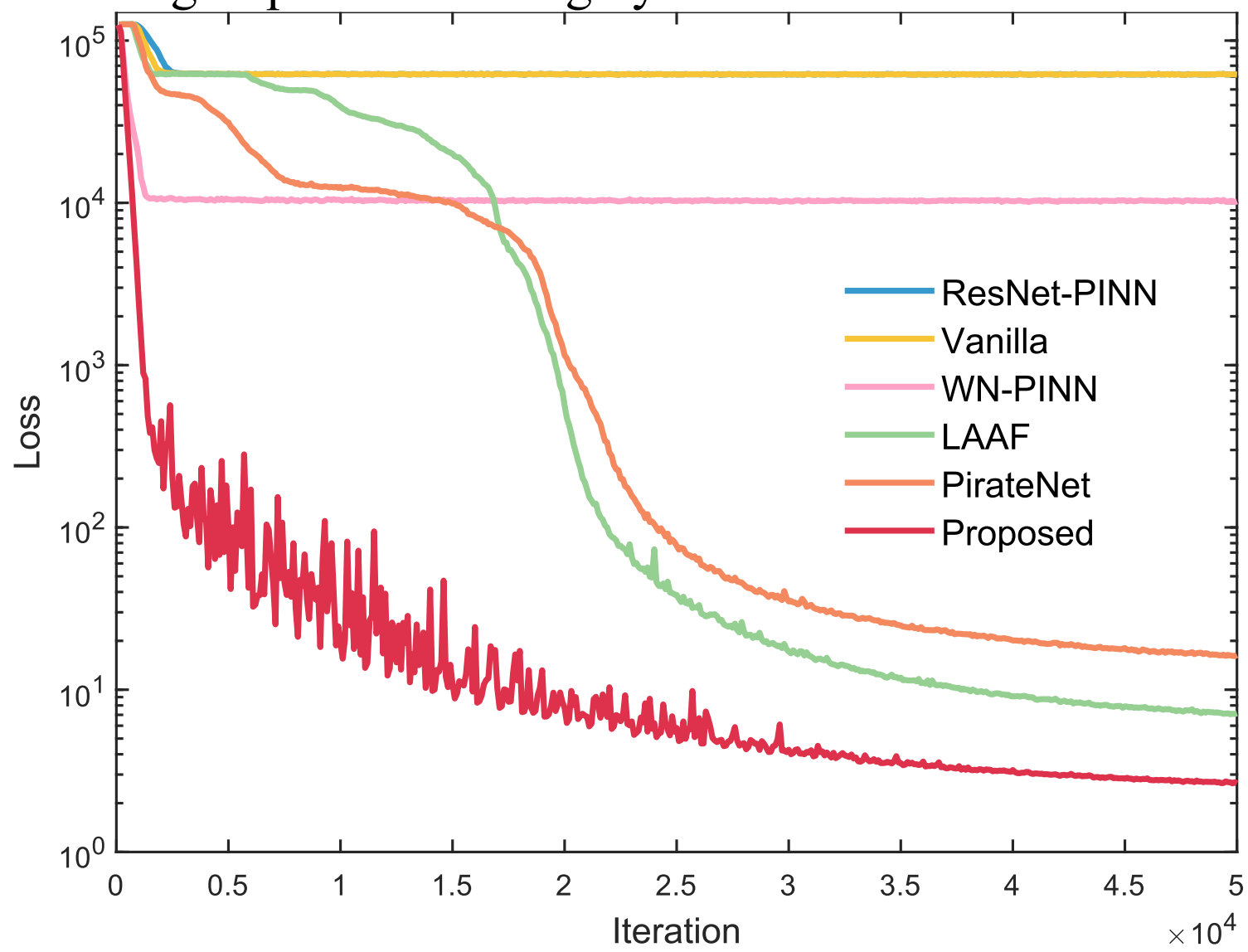

Figure 6. Helmholtz equation: averaged loss curves of different models with GELU as activation function.

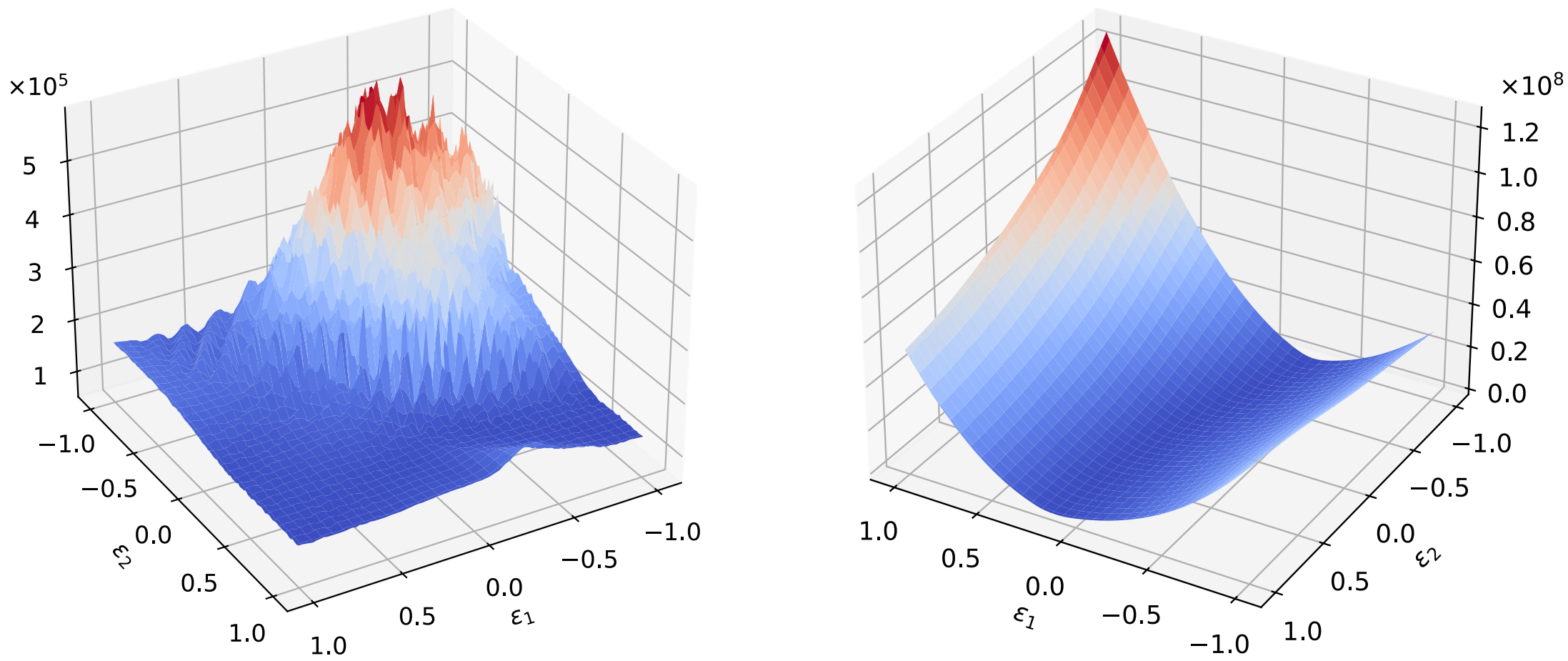

(a) Vanilla (b) Proposed

Figure 7. Helmholtz equation: loss landscapes of vanilla method and the proposed method with GELU as activation function.
To further explore the reason behind such difference, Fig. 7 presents the loss landscapes of vanilla method and the proposed method with GELU as activation function. We visualize the loss landscape by perturbing the trained model along the first two leading eigenvectors of the Hessian and evaluating the loss at each perturbed point. It's evident that the proposed method yields smoother loss landscape. For vanilla method, the surface is highly non-smooth, characterized by sharp spikes and irregular fluctuations. The irregular and rugged nature of the loss landscape indicates a tendency for the optimizer to become trapped in local minima with high loss value, which is consistent with the loss curve shown in Figure 6.

### 4.4 Wider Structure

Although it is well-known that deeper and wider neural networks can improve approximation capability, for PINNs the situation is quite different. Theoretical analysis and empirical experiments have proven that the PINNs degrade when the neural networks become wider or deeper [15]. While some solutions have been proposed to address the degradation problem in deep PINN architectures [15], [17], [34], the challenge of effectively utilizing wider architectures remains largely unexplored. Here we show that with the proposed method, PINNs can effectively utilize wider structure.

With the Convection equation mentioned above, we conduct experiments on neural networks with different widths across different activation functions. For all the neural networks, we fix the depth to 3 hidden layers. Other settings are aligned with section 4.1. All the results are averaged from 5 independent trials. As shown in Fig. 8, Vanilla PINNs suffer from a degradation issue as the width of the neural network increases. For the proposed method, a consistent improvement is witnessed as the neural network becomes wider. This indicates that the proposed method can mitigate the degradation problem observed in vanilla PINNs. Consequently, it allows PINNs to utilize wider structures, enhancing their robustness and accuracy.

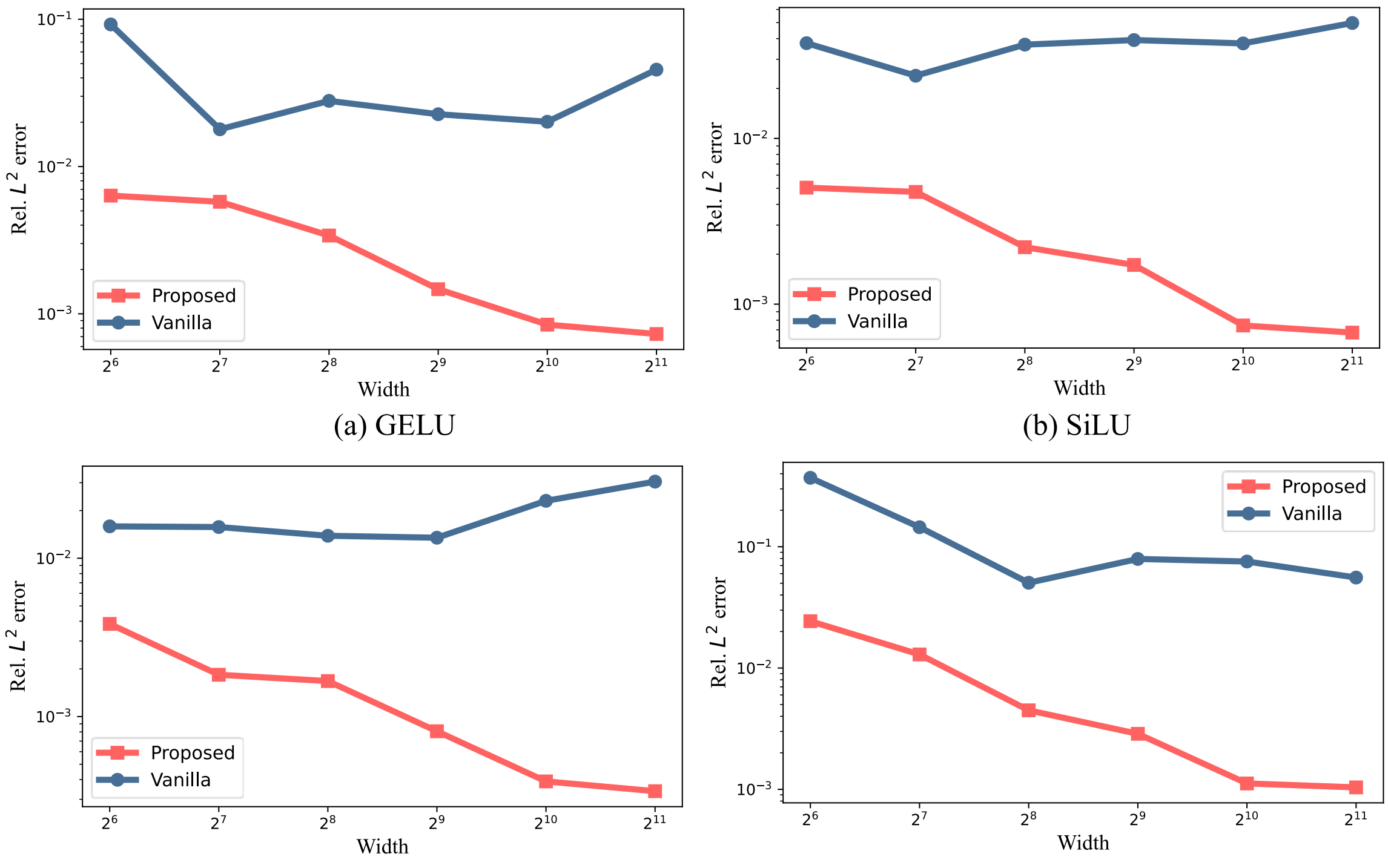

(a) GELU (b) SiLU

(c) Tanh (d) SoftPlus

Figure 8. Relative $L^2$ errors of vanilla PINNs and the proposed method across various network widths.

To further explore the underlying reasons behind this improvement, we present the pre-activation distributions of the vanilla and proposed methods in Fig. 9. It clearly shows that with the width increasing, the pre-activation distribution of vanilla PINNs tends to drift away from the origin. Also note that for the GELU, SiLU, and SoftPlus activation functions, the pre-activation distributions in the vanilla method exhibit a clear tendency to drift toward the negative range. This suggests that neurons may become inactive or saturated due to these activation functions' unsymmetric activation properties, which severely limit the network's expressive power. The proposed method can effectively retain the balanced distribution shape with a narrower variance.

The results presented above highlight that the feature distribution instability and neuron saturation issues of wider neural networks lead to the degradation of PINNs. Our approach directly mitigates these issues by maintaining stable pre-activation distributions, allowing wider architectures to maintain their representational capability and achieve better accuracy.

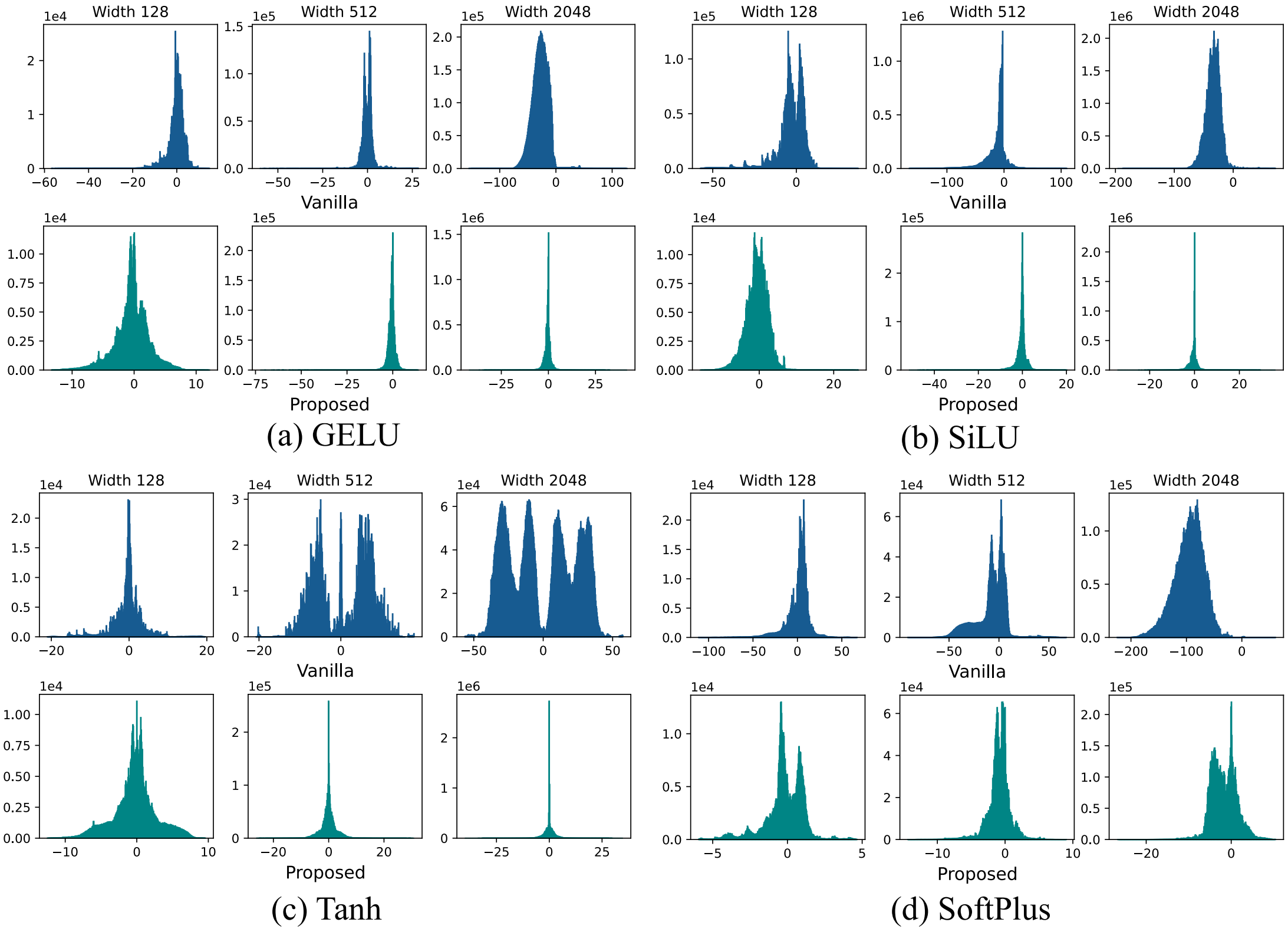

Figure 9. The pre-activation distributions of the 3rd hidden layer at the end of the training.

# 5 Neural Tangent Kernel Analysis

In this section, we investigate the training dynamics of Mask-PINN by its neural tangent kernel (NTK). Following the derivation of [32] , [33], we define the neural tangent kernel operator $\mathbf{K}$:

$$\mathbf{K}(\tau)=\begin{bmatrix}\mathbf{K}_{uu}(\tau) & \mathbf{K}_{ur}(\tau)\\ \mathbf{K}_{ru}(\tau) & \mathbf{K}_{rr}(\tau)\end{bmatrix}, \tag{44}$$

where $\mathbf{K}_{ru}(\tau)=\mathbf{K}_{ur}^{T}(\tau)$. Here $\tau$ represents iteration. The specific entries are given by:

$$(\mathbf{K}_{uu})_{ij}(\tau)=\left\langle \frac{d\mathcal{B}(x_b^i;\theta)}{d\theta}, \frac{d\mathcal{B}(x_b^j;\theta)}{d\theta} \right\rangle, \tag{45}$$

$$(\mathbf{K}_{ur})_{ij}(\tau)=\left\langle \frac{d\mathcal{B}(x_b^i;\theta)}{d\theta}, \frac{d\mathcal{N}\left[u(x_r^j;\theta)\right]}{d\theta} \right\rangle, \tag{46}$$

$$(\mathbf{K}_{rr})_{ij}(\tau)=\left\langle \frac{d\mathcal{N}\left[u(x_r^i;\theta)\right]}{d\theta}, \frac{d\mathcal{N}\left[u(x_r^j;\theta)\right]}{d\theta} \right\rangle. \tag{47}$$

NTK theory demonstrates that, with an infinitesimally small learning rate the gradient descent process can be represented as a continuous-time gradient flow:

$$\begin{bmatrix} \dfrac{d\mathcal{B}\left(x_b;\theta(\tau)\right)}{dt} \\ \dfrac{d\mathcal{N}\left[u\left(x_r;\theta(\tau)\right)\right]}{dt} \end{bmatrix} \approx -\mathbf{K}\cdot\begin{bmatrix} \mathcal{B}\left(x_b;\theta(\tau)\right)-g(x_b) \\ \mathcal{N}\left[u\left(x_r;\theta(\tau)\right)\right]-f(x_r) \end{bmatrix}, \tag{48}$$

$\theta(\tau)$ denotes the parameters of the network at $\tau$. The solution of the ODE can be expressed as:

$$\begin{bmatrix} \mathcal{B}\left(x_b;\theta(\tau)\right) \\ \mathcal{N}\left[u\left(x_r;\theta(\tau)\right)\right] \end{bmatrix} \approx \left(I-e^{-\mathbf{K}\tau}\right)\cdot\begin{bmatrix} g(x_b) \\ f(x_r) \end{bmatrix}, \tag{49}$$

Since $\mathbf{K}$ is positive semi-definite, we can apply its spectral decomposition as $\mathbf{K}=\mathbf{Q}\mathbf{\Lambda}\mathbf{Q}^T$, where $\boldsymbol{Q}$ is an orthogonal matrix whose $i$-th column is the eigenvector $q_i$ of $\mathbf{K}$. $\mathbf{\Lambda}$ is a diagonal matrix containing the corresponding eigenvalues. Using the identity $e^{-K\tau}=Q^T e^{-\Lambda\tau}Q$, we obtain:

$$\mathbf{Q}^T\begin{bmatrix} \mathcal{B}\left(x_b;\theta(\tau)\right)-g(x_b) \\ \mathcal{N}\left[u\left(x_r;\theta(\tau)\right)\right]-f(x_r) \end{bmatrix} \approx -e^{-\Lambda\tau}\mathbf{Q}^T\begin{bmatrix} g(x_b) \\ f(x_r) \end{bmatrix}, \tag{50}$$

The above equation indicates that the NTK's eigenvalues directly govern the convergence behaviour. Here we analyse the spectral properties of NTK under Burger's equation which expressed as follows:

$$u_t+uu_x-(0.01/\pi)u_{xx}=0,\ x\in[-1,1],\ t\in[0,1], \tag{51}$$

$$u(0,x)=-\sin(\pi x), \tag{52}$$

$$u(t,-1)=u(t,1)=0. \tag{53}$$

The comparison is conducted between the proposed method and vanilla PINN. Both have 5 hidden layers with 512 neurons each, and Tanh is adopted as activation function. The evolution of the NTK eigenvalue spectra is given in Fig. 10. Fig. 10 (a) reveals that the NTK spectra of the proposed method consistently outperforms vanilla across training. At small τ, it exhibits noticeably larger leading eigenvalues and a slower decay, which indicates richer feature representation and faster convergence. As τ increases, the gap narrows, but the proposed spectra still stay above vanilla across most of the range, showing it maintains a persistent advantage. In Fig. 10 (b), the proposed method's NTK spectrum converges to its steady shape significantly earlier than vanilla method, indicating stable optimization and efficient learning. These are confirmed by the loss and relative $L^2$ error curves observed during training. As shown in Fig. 11, both the curves of the proposed method decrease faster and to lower final values.

The NTK spectra reveal that Mask-PINNs maintain higher eigenvalues and converge to a steady shape earlier than the vanilla PINN, indicating more stable and efficient training dynamics.

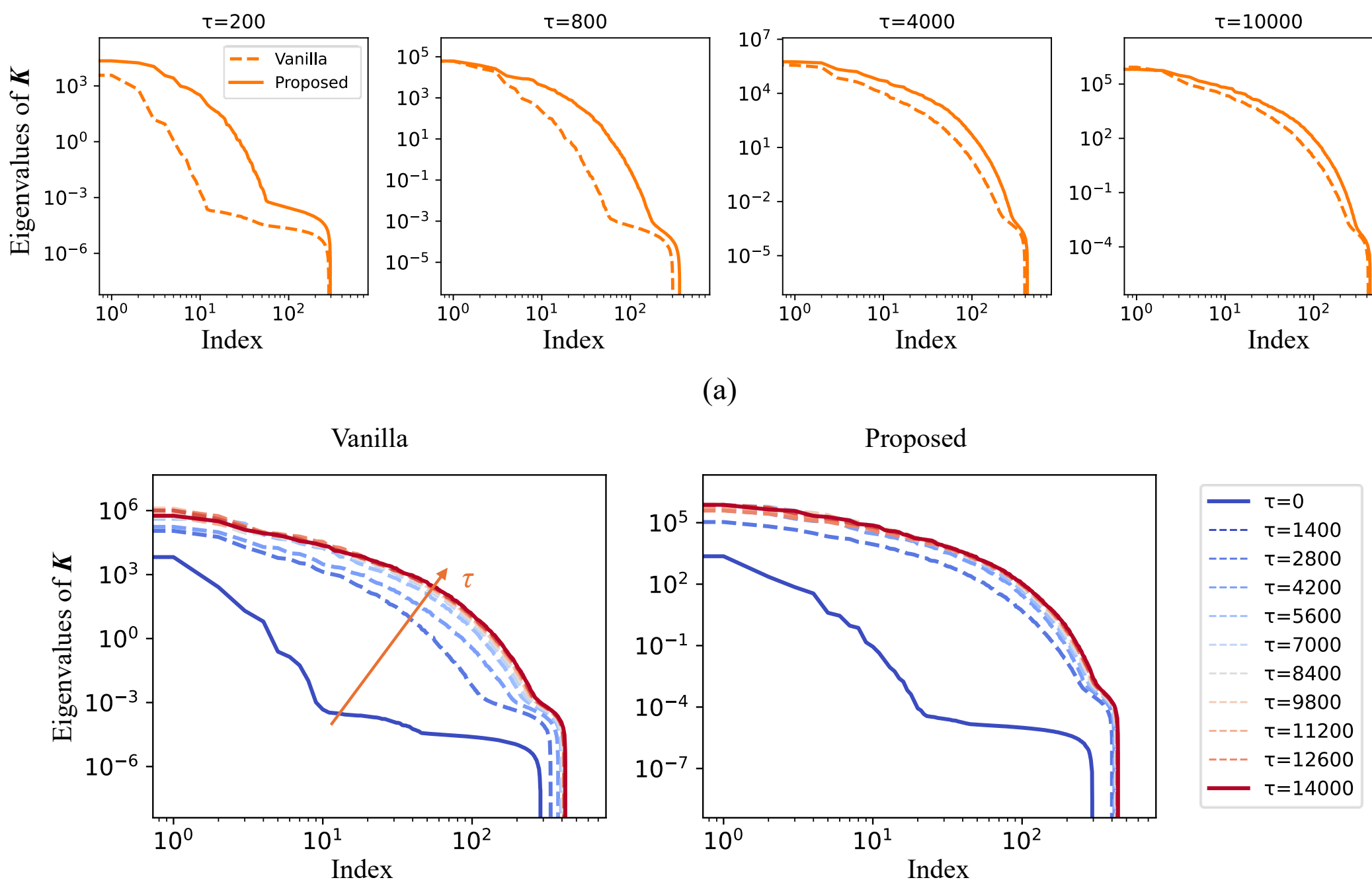

Figure 10. Evolution of the NTK eigenvalue spectra during training.

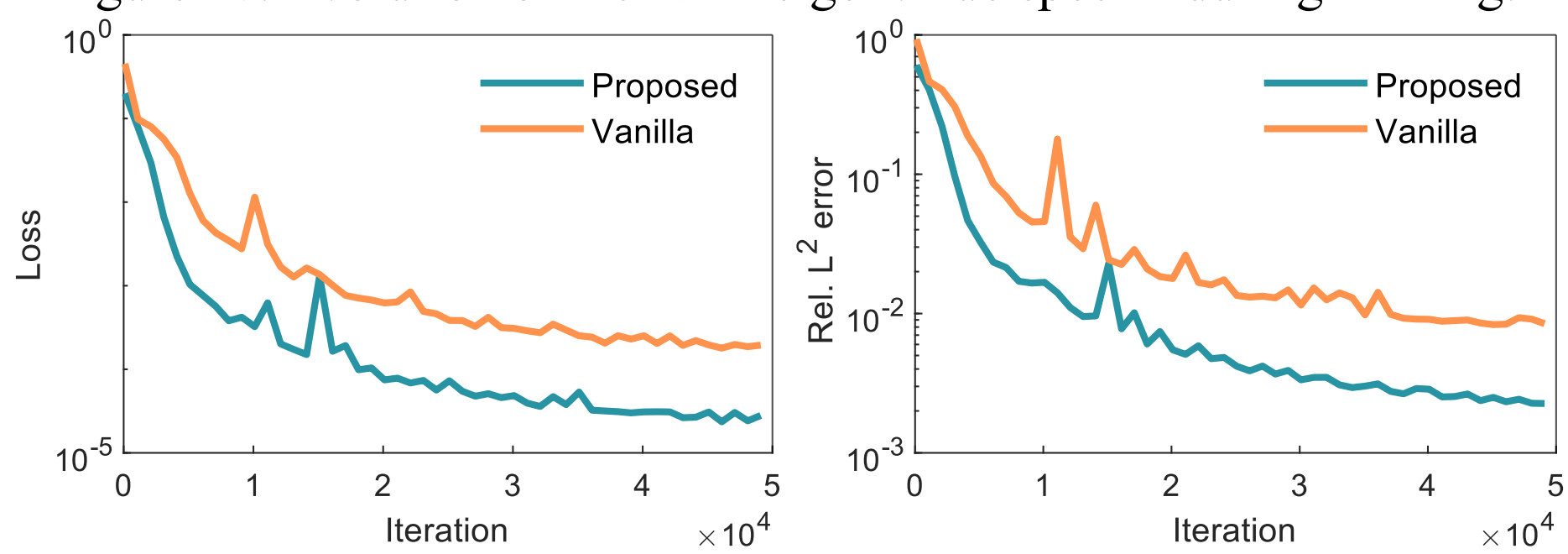

Figure 11. Loss and relative $L^2$ error curves during training.

# 6 Conclusion

In this work, we focus on internal covariate shift, a well-known problem in deep learning that remains unresolved in PINNs and significantly undermines training stability and solution accuracy. We proposed Mask-PINNs, a novel architecture designed to alleviate this issue of PINNs without compromising the physical constraints encoded in the model. Specifically, a smooth, learnable mask function is applied pointwise across hidden layers to regulate feature distributions thereby mitigating activation drift. This leads to significantly improved training stability, predictive accuracy, and robustness across a wide range of PDE benchmarks and activation functions. Our experiments demonstrate that Mask-PINNs not only outperform existing methods across different settings but also enable the effective use of wider network architectures. By stabilizing feature distributions, Mask-PINNs provide a simple yet powerful tool for enhancing the representational capability of PINNs. The introduction of Mask-PINNs represents a significant advancement towards stable and scalable neural network architectures for solving PDEs, opening several promising avenues for future exploration in PINNs.

Since the proposed mask function mitigates internal covariate shift while preserving the coordinate-based nature of PINNs, it is worth exploring its applicability to other coordinate-based tasks, such as implicit neural representations [35]. While our results are promising, further investigation is needed. Notably, we observed that the initial value of the mask scaling parameter α has a considerable impact on training dynamics. Future research should aim to develop principled initialization strategies for α, tailored to specific tasks and network configurations, to further enhance the reliability and performance of Mask-PINNs.

# References

[1] M. Raissi, P. Perdikaris, and G. E. Karniadakis, "Physics-informed neural networks: A deep learning framework for solving forward and inverse problems involving nonlinear partial differential equations," *Journal of Computational physics,* vol. 378, pp. 686-707, 2019.

[2] X. Jia *et al.*, "Physics guided RNNs for modeling dynamical systems: A case study in simulating lake temperature profiles," in *Proceedings of the 2019 SIAM international conference on data mining*, 2019: SIAM, pp. 558-566.

[3] M. Raissi, A. Yazdani, and G. E. Karniadakis, "Hidden fluid mechanics: Learning velocity and pressure fields from flow visualizations," *Science,* vol. 367, no. 6481, pp. 1026-1030, 2020.

[4] S. Cai *et al.*, "Artificial intelligence velocimetry and microaneurysm-on-a-chip for three-dimensional analysis of blood flow in physiology and disease," *Proceedings of the National Academy of Sciences,* vol. 118, no. 13, p. e2100697118, 2021.

[5] J. Zhang and X. Zhao, "Three-dimensional spatiotemporal wind field reconstruction based on physics-informed deep learning," *Applied Energy,* vol. 300, p. 117390, 2021.

[6] S. Cai, Z. Mao, Z. Wang, M. Yin, and G. E. Karniadakis, "Physics-informed neural networks (PINNs) for fluid mechanics: A review," *Acta Mechanica Sinica,* vol. 37, no. 12, pp. 1727-1738, 2021.

[7] S. Cai, Z. Wang, S. Wang, P. Perdikaris, and G. E. Karniadakis, "Physics-informed neural networks for heat transfer problems," *Journal of Heat Transfer,* vol. 143, no. 6, p. 060801, 2021.

[8] H. Hu, L. Qi, and X. Chao, "Physics-informed Neural Networks (PINN) for computational solid mechanics: Numerical frameworks and applications," *Thin-Walled Structures,* p. 112495, 2024.

[9] X.-X. Chen, P. Zhang, and Z.-Y. Yin, "Physics-Informed neural network solver for numerical analysis in geoengineering," *Georisk: Assessment and Management of Risk for Engineered Systems and Geohazards,* vol. 18, no. 1, pp. 33-51, 2024.

[10] Z. Wu, H. Wang, C. He, B. Zhang, T. Xu, and Q. Chen, "The application of physics-informed machine learning in multiphysics modeling in chemical engineering," *Industrial & Engineering Chemistry Research,* vol. 62, no. 44, pp. 18178-18204, 2023.

[11] Y. Bai, T. Chaolu, and S. Bilige, "The application of improved physics-informed neural network (IPINN) method in finance," *Nonlinear Dynamics,* vol. 107, no. 4, pp. 3655-3667, 2022.

[12] K. Kashinath *et al.*, "Physics-informed machine learning: case studies for weather and climate modelling," *Philosophical Transactions of the Royal Society A,* vol. 379, no. 2194, p. 20200093, 2021.

[13] H. Wang, L. Lu, and G. Huang, "Learning Specialized Activation Functions for Physics-Informed Neural Networks," *Communications in Computational Physics,* vol. 34, no. 4, pp. 869-906, 2023.

[14] J. Zhang and C. Ding, "Simple yet effective adaptive activation functions for physics-informed neural networks," *Computer Physics Communications,* vol. 307, p. 109428, 2025.

[15] S. Wang, B. Li, Y. Chen, and P. Perdikaris, "PirateNets: Physics-informed Deep Learning with Residual Adaptive Networks," *arXiv preprint arXiv:2402.00326,* 2024.

[16] M. Cooley, R. M. Kirby, S. Zhe, and V. Shankar, "HyResPINNs: Adaptive Hybrid Residual Networks for Learning Optimal Combinations of Neural and RBF Components for Physics-Informed Modeling," *arXiv preprint arXiv:2410.03573,* 2024.
[17] F. Jiang, X. Hou, and M. Xia, "Element-wise multiplication based deeper physics-informed neural networks," *arXiv preprint arXiv:2406.04170,* 2024.
[18] V. Sitzmann, J. Martel, A. Bergman, D. Lindell, and G. Wetzstein, "Implicit neural representations with periodic activation functions," *Advances in neural information processing systems,* vol. 33, pp. 7462-7473, 2020.
[19] H. woo Lee, H. Choi, and H. Kim, "Robust Weight Initialization for Tanh Neural Networks with Fixed Point Analysis," in *The Thirteenth International Conference on Learning Representations*.
[20] R. Bischof and M. A. Kraus, "Multi-objective loss balancing for physics-informed deep learning," *Computer Methods in Applied Mechanics and Engineering,* vol. 439, p. 117914, 2025.
[21] Q. Liu, M. Chu, and N. Thuerey, "Config: Towards conflict-free training of physics informed neural networks," *arXiv preprint arXiv:2408.11104,* 2024.
[22] S. Wang, A. K. Bhartari, B. Li, and P. Perdikaris, "Gradient Alignment in Physics-informed Neural Networks: A Second-Order Optimization Perspective," *arXiv preprint arXiv:2502.00604,* 2025.
[23] S. Wang, Y. Teng, and P. Perdikaris, "Understanding and mitigating gradient flow pathologies in physics-informed neural networks," *SIAM Journal on Scientific Computing,* vol. 43, no. 5, pp. A3055-A3081, 2021.
[24] H. Gao, L. Sun, and J.-X. Wang, "PhyGeoNet: Physics-informed geometry-adaptive convolutional neural networks for solving parameterized steady-state PDEs on irregular domain," *Journal of Computational Physics,* vol. 428, p. 110079, 2021.
[25] K. Shukla, J. D. Toscano, Z. Wang, Z. Zou, and G. E. Karniadakis, "A comprehensive and FAIR comparison between MLP and KAN representations for differential equations and operator networks," *Computer Methods in Applied Mechanics and Engineering,* vol. 431, p. 117290, 2024.
[26] B. Yuan, H. Wang, A. Heitor, and X. Chen, "f-PICNN: a physics-informed convolutional neural network for partial differential equations with space-time domain," *Journal of Computational Physics,* vol. 515, p. 113284, 2024.
[27] S. Ioffe and C. Szegedy, "Batch normalization: accelerating deep network training by reducing internal covariate shift," presented at the Proceedings of the 32nd International Conference on International Conference on Machine Learning - Volume 37, Lille, France, 2015.
[28] C. Nwankpa, W. Ijomah, A. Gachagan, and S. Marshall, "Activation functions: Comparison of trends in practice and research for deep learning," *arXiv preprint arXiv:1811.03378,* 2018.
[29] M. Lee, "Mathematical analysis and performance evaluation of the gelu activation function in deep learning," *Journal of Mathematics,* vol. 2023, no. 1, p. 4229924, 2023.
[30] S. R. Dubey, S. K. Singh, and B. B. Chaudhuri, "Activation functions in deep learning: A comprehensive survey and benchmark," *Neurocomputing,* vol. 503, pp. 92-108, 2022.
[31] J. L. Ba, "Layer normalization," *arXiv preprint arXiv:1607.06450,* 2016.
[32] T. Salimans and D. P. Kingma, "Weight normalization: A simple reparameterization to accelerate training of deep neural networks," *Advances in neural information processing systems,* vol. 29, 2016.

[33] A. D. Jagtap, K. Kawaguchi, and G. Em Karniadakis, "Locally adaptive activation functions with slope recovery for deep and physics-informed neural networks," *Proceedings of the Royal Society A,* vol. 476, no. 2239, p. 20200334, 2020.
[34] A. Aygun, R. Maulik, and A. Karakus, "Physics-informed neural networks for mesh deformation with exact boundary enforcement," *Engineering Applications of Artificial Intelligence,* vol. 125, p. 106660, 2023.
[35] J. J. Park, P. Florence, J. Straub, R. Newcombe, and S. Lovegrove, "Deepsdf: Learning continuous signed distance functions for shape representation," in *Proceedings of the IEEE/CVF conference on computer vision and pattern recognition*, 2019, pp. 165-174.